\documentclass{article}

\usepackage[nonatbib,final]{nips_2018}

\usepackage[utf8]{inputenc} 
\usepackage[T1]{fontenc}    
\usepackage{hyperref}       
\usepackage{url}            
\usepackage{booktabs}       
\usepackage{amsfonts}       
\usepackage{nicefrac}       
\usepackage{microtype}      
\usepackage{graphicx}
\usepackage{float}
\usepackage{stfloats}
\usepackage{multirow}
\graphicspath{{./figures/}}
\usepackage{color}
\usepackage{caption}
\title{Multivariate Time-series Similarity Assessment via Unsupervised Representation Learning and Stratified Locality Sensitive Hashing: Application to Early Acute Hypotensive Episode Detection}
\author{
  Jwala~Dhamala$^1$ \ Emmanuel~Azuh$^2$ \ Abdullah~Al-Dujaili$^2$ \ Jonathan~Rubin$^3$ \ Una-May~O'Reilly$^2$ 
  \\
  $^1$Rochester Institute of Technology \ $^2$Massachusetts Institute of Technology \ $^3$Philips Research North America 
}

\begin{document}
\newcommand{\jd}[1]{\color{blue}#1 \color{black}}
\newcommand{\jr}[1]{\color{red}#1 \color{black}}

\maketitle
\vspace{-.2cm}
\begin{abstract}
Timely prediction of clinically critical events in Intensive Care Unit (ICU) is important for improving care and survival rate. Most of the existing approaches are based on the application of various classification methods on explicitly extracted statistical features from vital signals. 
In this work, we propose to eliminate the high cost of engineering hand-crafted features from multivariate time-series of physiologic signals by learning their representation with a sequence-to-sequence auto-encoder. We then propose to hash the learned representations 
to enable signal similarity assessment for the prediction of critical events. We apply this methodological framework to predict Acute Hypotensive Episodes (AHE) on a large and diverse dataset of vital signal recordings. 
Experiments demonstrate the ability of the presented framework in accurately predicting an upcoming AHE. 
\end{abstract}
\vspace{-.2cm}
\section{Introduction}
\vspace{-.1cm}
Acute Hypotensive Episodes (AHE) among patients in the Intensive Care Unit (ICU) can lead to multiple organ failure and even death~\cite{moody2009predicting,zenati2002brief}. An accurate and early detection of such clinically critical event is important for improving treatment and survival rate among ICU patients.

In the past decade numerous efforts have been made at early AHE detection. Most of these efforts are machine learning approaches in which statistical features from various vital signals of ICU patients are first extracted and then utilized within a classification framework. For example, in \cite{rocha2011prediction} samples from Arterial Blood Pressure (ABP) signal are utilized in a Neural Network, in \cite{chen2009forecasting} multiple features such as an average and exponential weighted average from both the ABP vital signal and waveform are utilized in a classification framework, and in \cite{de2017distributed} averages of the ABP signal over every five minutes are utilized in Distributed Stratified Locality Sensitive Hashing. 
Limited efforts have been made to incorporate information from multiple vital signals~\cite{chen2009forecasting}. The majority of existing works utilize features extracted from the ABP signal and Mean Arterial Pressure (MAP) as their fundamental measurement. However, by utilizing a single signal and MAP, these approaches fail to incorporate the different correlations that could exist among the various vital signals. Moreover, all of the existing methods rely on explicitly extracting features from a few selected vital signals. However, none of these hand-crafted signal features have the ability to capture the complex time oriented patterns that exist within one signal, and, with delays and time dilations, across different signals. 
\begin{figure*}[!t]
	\centering 
    \includegraphics[width=0.8\textwidth]{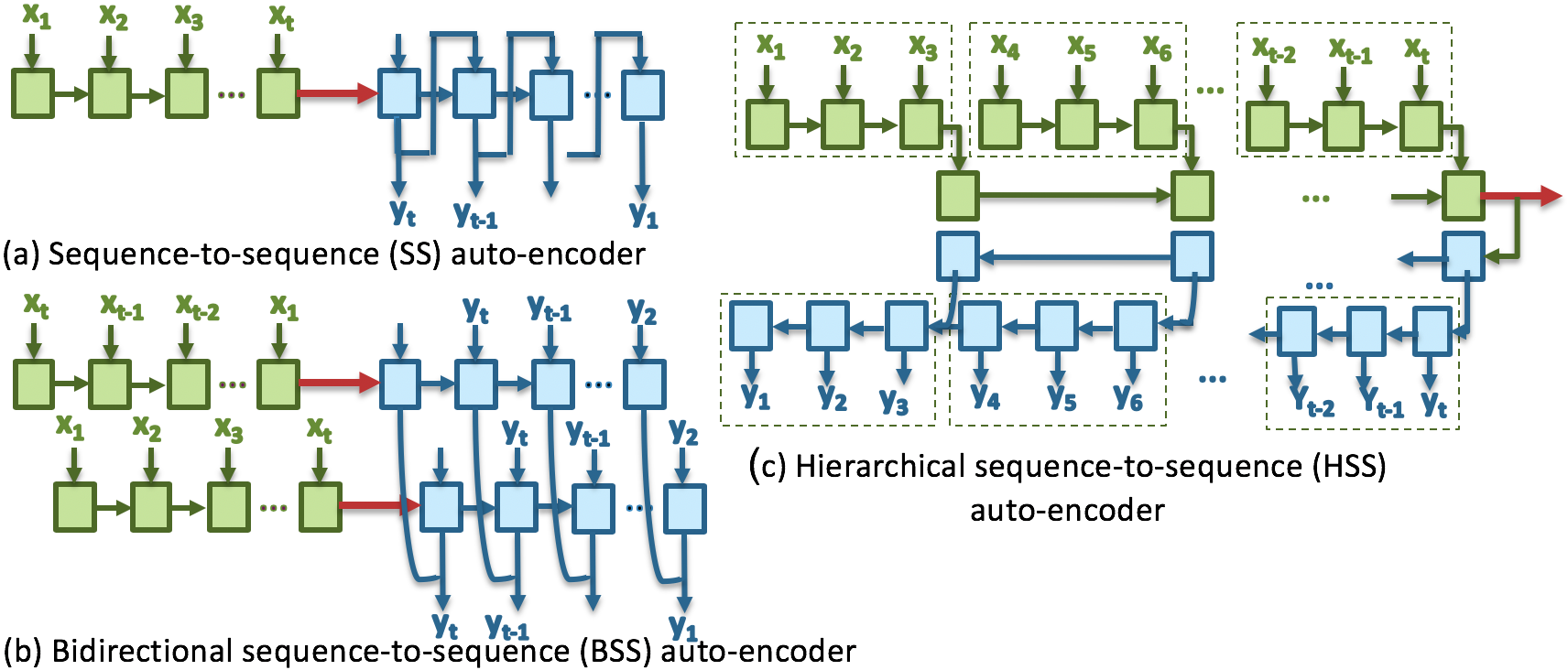}
    \caption{\small{Different sequence-to-sequence auto-encoder models explored in this paper. Each box represents a GRU. Red arrow represents the vectorial representation (context vector) of multivariate time-series signals.}\vspace{-.4cm}}
    \label{fig:architectures}
\end{figure*}

In this study, we present a framework that goes beyond explicit hand-crafted feature based signal similarity assessment. It is achieved by unsupervised learning of representations from a large training dataset of multivariate time-series ICU signals using a sequence-to-sequence gated recurrent unit (GRU) based auto-encoder; the learning model will capture the complex temporal dynamics within a signal and the inter-dependency among various signals in a fixed-length vector embedding. These compact and information rich vectorial representations of the training multivariate signals are used to hash a labeled dataset to support similarity-based retrieval. The similarity set then supports prediction. It also provides clinicians with historical comparisons. 

The presented framework is applied to AHE prediction on a large and diverse dataset containing 16 vitals signal recordings from over 20,000 different patients ICU stays at 10 different ICU centers. 
 Our preliminary study achieves an accuracy of 75.07$\%$ in early AHE detection showing the feasibility of this method. To the best of our knowledge this is the first attempt at utilizing a large database of multivariate vital signals for AHE prediction. We further study the learned representations to interpret how information from various signals are encoded within them. This framework provides a general solution to predicting critical ICU events 
 beyond the tested application of early AHE detection.
\vspace{-.2cm}
\section{Unsupervised Learning of Multivariate Vital Signal Representations}
\label{sec:unsupervised}
\vspace{-.1cm}
Recently auto-encoders have shown the ability to learn meaningful representations of data in various domains~\cite{li2015hierarchical,dhamala2018high,schuster1997bidirectional}. Here, we utilize it to learn the representations of multivariate vital signals.

An auto-encoder consists of two multi-layered Recurrent Neural Networks (RNN): an encoder RNN that summarizes the multivariate time-series of vital signals into a fixed-length vector and a decoder RNN that is able to reconstruct the original signals from this vector alone. In this study, we use RNN with gated recurrent units (GRUs) modules~\cite{chung2014empirical}. More specifically, given a multivariate time-series signal $\textbf{X}^{(i)} =  \{\textbf{x}_1^{(i)}, \textbf{x}_2^{(i)}, ..., \textbf{x}_{T}^{(i)} \}$, the encoder RNN sequentially takes as inputs the current input sample $\textbf{x}_t$ and the previous hidden state $\textbf{h}_{t-1}$ and outputs a vector of hidden state $\textbf{h}_t$. 
The hidden state at the final time instant 
$\textbf{h}_{T}$ (termed as the context vector) embodies all the crucial characteristics from the entire  
signals sequence.  The decoder RNN takes this context vector $\textbf{h}_{T}$ and sequentially reconstructs the vital signals as $\textbf{Y}^{(i)} =  \{\textbf{y}_T^{(i)}, \textbf{y}_{T-1}^{(i)}, ... \textbf{y}_{1}^{(i)} \}$ taking at each time step as inputs the hidden state and the output from the previous time step. A reconstruction error consisting of the sum of the squared error between the input $\textbf{X}$ and the reconstructed $\textbf{Y}$ signals is minimized to train the network. The architecture of a standard sequence-to-sequence auto-encoder (SS) is shown in Fig.~\ref{fig:architectures}a. Because the burden to summarize all the signal's information is entirely upon the context vector, we experimented with two different improvements over the baseline SS model.\\
\textbf{Bidirectional sequence-to-sequence auto-encoder (BSS):} 
The context vector capturing the progression of vital signals from the start time to the end time can learn different information 
than that which captures the progression of these signals in reverse direction. To allow incorporation of both 
we replace the encoder in SS model with a bidirectional encoder as shown in Fig.~\ref{fig:architectures}b~\cite{schuster1997bidirectional}. Here, the decoder must reconstruct the multivariate vital signals with the final hidden states from both the encoders. The final context vector is the concatenation of the two final hidden states.\\
\textbf{Hierarchical sequence-to-sequence auto-encoder (HSS):} Although GRUs are known to be capable of capturing long-range dependencies, the information 
at the beginning of the sequence might not be strongly encoded in the final context vector.  To attempt to capture dense information from throughout the sequence length we utilize a hierarchical model~\cite{li2015hierarchical}. As shown in Fig.~\ref{fig:architectures}c, this model consists of two layers of encoders in which the first layer encodes $n$ sections of the entire sequence into $n$ sub-contexts and the second layer encodes these $n$ sub-contexts into a final context vector. We also explore a model that combines BSS and HSS models termed as the bidirectional hierarchical 
model (BHSS). In this model, the encoders in the HSS are replaced with bidirectional encoders.
\vspace{-.2cm}
\section{Similarity Assessment with Stratified Locality Sensitive Hashing}
\label{ssection:lsh}
\vspace{-.1cm}
Approximate nearest neighbor based approaches 
have shown to enable efficient similarity-based retrieval from massive scale data~\cite{unastratified}. However, because these methods require fixed-length vectors as features, they cannot be directly applied for the similarity assessment of multivariate time-series signals. In this work, we generalize the features for arbitrary multivariate signals by encoding them into fixed-length vectors as described in Section~\ref{sec:unsupervised}. These vectorial representations then enable hashing-based multivariate time-series similarity assessment.

For an accurate and efficient similarity-based retrieval of vital signals representations, we use an approximate nearest neighbor technique called 
Stratified Locality Sensitive Hashing (SLSH)~\cite{unastratified}. SLSH consists of two stages: table construction and query resolution. During table construction, the context vectors of a dataset $\mathcal{D}$ are first split into different buckets in outer tables using locality sensitive hash functions. $(r,cr,p_1,p_2)$ - sensitive hash functions for a metric $\delta(.,.)$ are hash functions belonging to a family $\mathcal{H} = h : \mathcal{X} \rightarrow \mathcal{U}$ such that, for any two context vectors $\textbf{x},\textbf{y} \in \mathcal{X}$ and two probabilities $p_1 > p_2$: if $\delta(\textbf{x},\textbf{y})\leq r$ then $P_{\mathcal{H}}[h(\textbf{x}) = h(\textbf{y})] \geq p_1$, and if $\delta(\textbf{x},\textbf{y})\geq cr$ then $P_\mathcal{H}[h(\textbf{x})=h(\textbf{y})]\leq p_2$, where $\mathcal{U}$ is the hash space~\cite{indyk1998approximate}. We use two hash families: the bit-sampling family in which $\delta$ is l1 norm and the Random Projection family in which $\delta$ is the Cosine norm~\cite{unastratified}. To reduce false positives, a hash family $\mathcal{H}'$ made by the concatenation of $m_{out}$ independent hash functions uniformly sampled from $\mathcal{H}$ is used. To reduce false negatives, $L_{out}$ independent tables indexed by $\mathcal{H}'$ are used. Furthermore, for each bucket in the $L_{out}$ outer tables that has more than $\alpha N$ context vectors where $N$ is the size of dataset, the $\alpha N$ context vectors in this bucket are mapped to $L_{in}$ inner tables with a different hash family formed by concatenating $m_{in}$ hash functions. The splitting into inner layer will improve the efficiency by reducing the number of candidates during query stage and improve the accuracy by enabling to incorporate a different similarity metric through hashing function from a different family. In the query stage, a set of candidates consisting of the union of context vectors hashed to the same bucket in each outer table or inner table (if exists) is chosen. On these candidates linear scanning is done to pick the nearest neighbor. 
\vspace{-.2cm}
\section{Experiments}
\vspace{-.1cm}
\paragraph{Dataset extraction and pre-processing:}
We extracted data from the large multi-center critical care database: eICU-CRD~\cite{PhysioNet,PollardSD2018}. It consists of 16 vital signals of over 20,000 patients stays in 10 ICU centers. The vital signal's samples are five minutes median of one minute averages. 
Following the definition of AHE in \cite{moody2009predicting}, we first identified signal segments that either contain AHE (MAP $\leq$ 60mmHg for 30 minutes) or not. For unsupervised representation learning, we extracted vital signals of length six hours 
that precede a lead time of 30 minutes and the identified segments (both with and without AHE). The lead time ensures early detection of AHE. For AHE classification with SLSH, the multivariate vital signals are labeled as positive if the following MAP signal segment had AHE else, as negative. 
We excluded two vital signals that were missing more than 85$\%$ of their samples. However, we included six vital signals that were missing more than $\sim50\%$ of their samples. To handle these missing samples 
we used interpolation, forward fill, and backward fill. A vital signal that is entirely absent is imputed with the population mean of the corresponding vital sign. Finally, we balance the number of positive and negative examples by down-sampling. In this way, we extracted a total of 63,232 examples, each comprising 72 samples of 14 vital signals. After pre-processing, the dataset is feature normalized and split into train, validation, and test set in the ratio of 81:9:10. Training is done for 1000 epochs with the Adam optimizer~\cite{kingma2014adam} and a learning rate of 0.0001. We use SLSH with following parameter values: $L_{out}=96$, $m_{out}=100$, $L_{in}=30$, $m_{in}=10$, $\alpha=0.1$, and $k=1$~\cite{unastratified}. The accuracy of AHE prediction is measured in terms of two metrics: Accuracy = $\frac{(\mathrm{TP} + \mathrm{TN})}{\mathrm{TP} + \mathrm{TN} + \mathrm{FP} + \mathrm{FN}}$  and Matthews Correlation Coefficient $(\mathrm{MCC}) = \frac{\mathrm{TP} X \mathrm{TN} - \mathrm{FP} X \mathrm{FN}} { \sqrt{(\mathrm{TP} + \mathrm{FP})(\mathrm{TP}+\mathrm{FN})(\mathrm{TN}+\mathrm{FP})(\mathrm{TN}+\mathrm{FN})}}$, where TP, TN, FP, and FN respectively represent true positives, true negatives, false positives, and false negatives.

\begin{figure}[!t]
\begin{minipage}[!t]{\textwidth}
\centering
\begin{minipage}[!t]{0.6\textwidth}
\centering
\small
\begin{tabular}{lrrrr}
\hline
\multirow{2}{*}{\textbf{Model}} & \multicolumn{2}{c}{\textbf{Accuracy}} & \multicolumn{2}{c}{\textbf{MCC}} \\ \cline{2-5} 
& \multicolumn{1}{c}{\textbf{Val}} & \multicolumn{1}{c}{\textbf{Test}} & \multicolumn{1}{c}{\textbf{Val}} & \multicolumn{1}{c}{\textbf{Test}} \\ 
\hline 
Average MAP                & 0.6889  & 0.7054  & 0.3778     & 0.4106\\
Average of vital signals    & 0.7205  & 0.7325  & 0.4410     & 0.4652\\ 
\hline 
SS                          & 0.7262   & 0.7332                          & 0.4566    & 0.4727\\ 
HSS [9 samples]             & 0.7356   & 0.7329                          & 0.4745    & 0.4713\\
HSS [12 samples]            & 0.7357   & 0.7456                          & 0.4746    & 0.4960\\
BSS                         & 0.7480   & \textbf{0.7486}                 & 0.5007    & \textbf{0.5027}\\
BSS [2 layers]              & 0.7333   & 0.7355                          & 0.4727    & 0.4790\\
BHSS [9 samples]            & 0.7411   & {{\textbf{0.7506}}}   & 0.4875    & { {\textbf{0.5078}}}\\
BHSS [12 samples]           & 0.7416   & \textbf{0.7460}                 & 0.4875    & \textbf{0.4981} \\ \hline
\label{table:accuracy}
\end{tabular}
\captionof{table}{\small{Results of AHE prediction in validation (column: val) and test (column: test) datasets in terms of accuracy and MCC. Lead time of 30 minutes is used. Top three test scores are highlighted.}\vspace{-.3cm}}
\end{minipage}
\hfill
\begin{minipage}[!t]{0.35\textwidth}
\centering
\includegraphics[width=1\textwidth]{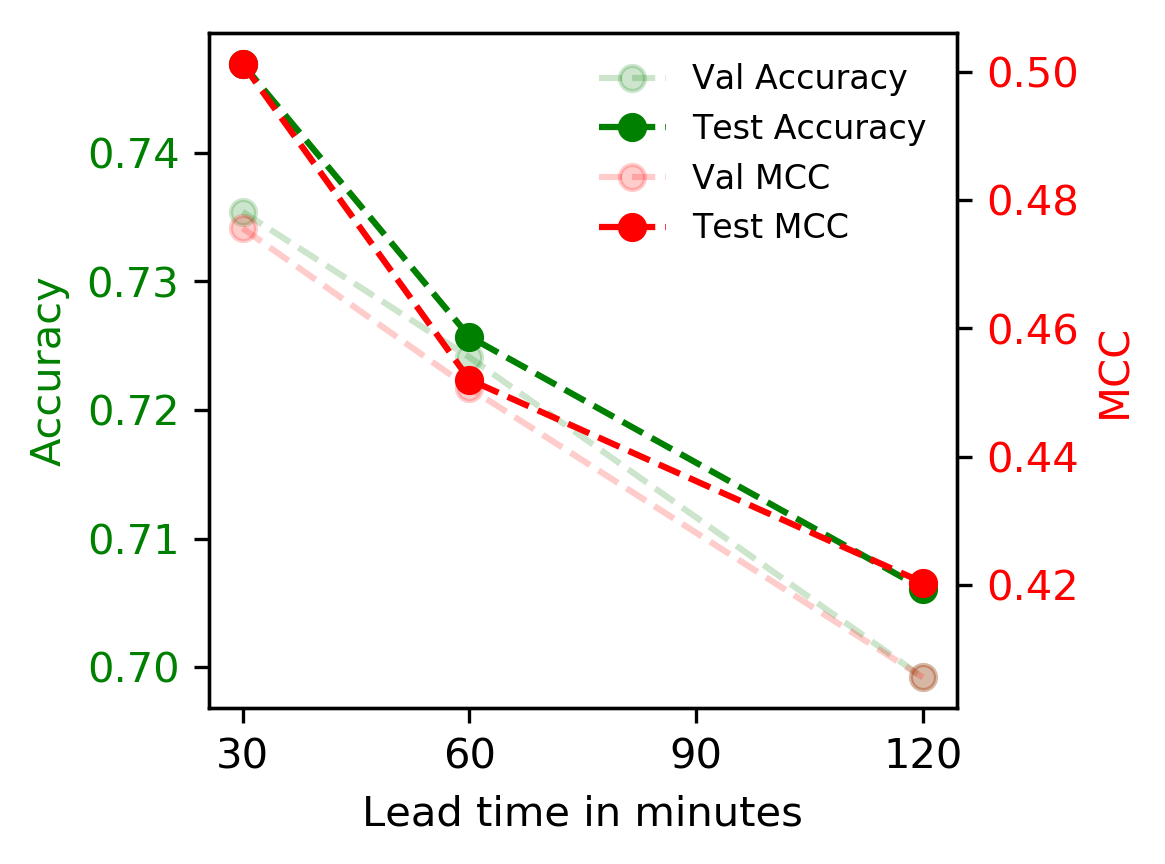}
\captionof{figure}{\small{Accuracy and MCC decrease as lead time increases in both validation and test dataset.}\vspace{-.3cm}}
\label{fig:leadtimes}
\end{minipage}
\end{minipage}
\end{figure}
\begin{figure*}[!t]
	\centering 
    \includegraphics[width=0.650\textwidth]{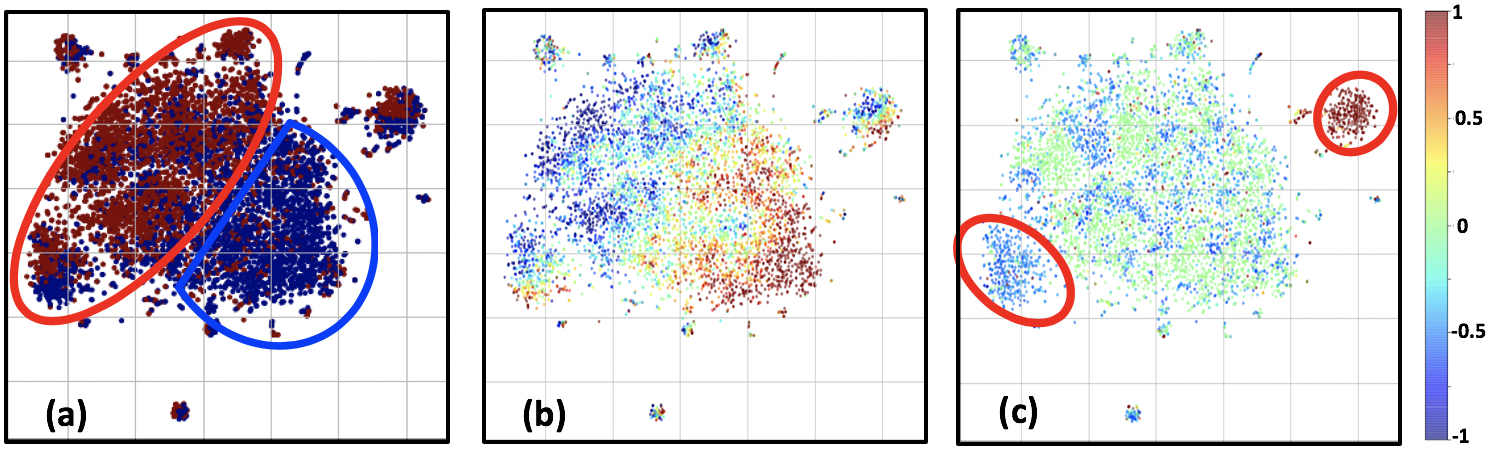}
    \caption{\small{tSNE on context vectors of test dataset from BSS model colored by (a) red: positive examples and blue: negative examples, (b) average systemic diastolic blood pressure; and (c) average central venous pressure.}\vspace{-.4cm}}
    \label{fig:context}
\end{figure*}
\textbf{Evaluation and Comparison:} We first evaluate the ability of vital signal representations learned with SS, BSS, HSS, and BHSS models in AHE prediction. Table 1 summarizes the accuracy with various models. 
All models show an improvement in accuracy over the SS model. Among all models, the BHSS with a sub-sequence of size nine shows the best performance (accuracy: 75.06 $\%$; MCC: $0.5078$). There is no drastic gain by using complex models. This is likely because the dataset has a large amount of missing values filled with population mean; simpler models are capable of capturing most of the information from this dataset. 

We also compare the performance of presented framework with two common approaches for AHE detection: 1) SLSH on the average of MAP as feature, and 2) SLSH on averages of all vital signals as features~\cite{langley2009predicting,chen2009forecasting}. Because the dataset used here is entirely new, we cannot entirely replicate the methods in the literature. The method using averages from all vital signals shows higher accuracy than that using the average from MAP alone showing that additional vital signals could improve the result. In comparison, the presented framework shows higher accuracy (see Table 1).

To study the ability of the presented framework in early AHE detection we experimented with three lead times: 30 minutes, one hour, and two hours. The result as shown in Fig.~\ref{fig:leadtimes} shows that the features learned from vital signals that are closer to the AHE episodes have higher accuracy, \textit{i.e.}, highest accuracy was attained with a lead time of 30 minutes and lowest with a lead time of two hours.

\textbf{Interpretation of  context vectors:}
To interpret the learned representations of vital signals, we apply t-Distributed Stochastic Neighbor Embedding (t-SNE) on context vectors of the test dataset obtained with the BSS model and visualize them~\cite{maaten2008visualizing}. As shown in Fig.~\ref{fig:context}a, positive and negative samples differentiate into separate clusters. A closer inspection by coloring them with the mean of the systemic diastolic blood pressure (DBP) (Fig.~\ref{fig:context}b) shows that from the regions of higher density to lower density of negative samples (Fig.\ref{fig:context}a blue to red) the mean DBP is gradually increasing indicating that negative samples, in general, have higher value of mean DBP which is captured by the learned representations. Similarly, coloring them with the mean of the central venous pressure signal (Fig.~\ref{fig:context}c) shows two small cluster: one with higher value and next with lower value. Throughout the largest cluster, values close to zeros are present. These samples represent the missing values that were imputed with population mean. This shows that the auto-encoder models have the ability to encode informative knowledge from 
that are general to predicting various ICU conditions. However, they can also learn uninformative knowledge, \textit{e.g.}, imputed data could negatively impact prediction results.
\vspace{-.2cm}
\section{Conclusion}
\vspace{-.1cm}
We presented a novel framework for multivariate time-series similarity assessment that is achieved by learning compact representations of multivariate time-series signals with a sequence-to-sequence auto-encoder and then using these implicitly learned features with SLSH for signal similarity assessment. Our preliminary experiments on early AHE prediction 
shows the feasibility of this approach. In the future, we will investigate improved methods to handle missing data~\cite{lipton2016modeling}, split of dataset based on patients, and 
auto-encoder models with higher attention on relevant features and time steps for an increased accuracy.

\bibliographystyle{plain}
\bibliography{ml4h_nips_2018.bib}

\end{document}